\documentclass[letterpaper, 10 pt, journal, twoside]{ieeetran}

\IEEEoverridecommandlockouts                              

\usepackage{amssymb}  
\usepackage{titlesec}

\usepackage{comment}

\usepackage{glossaries}
\makeglossaries

\newacronym{lfd}{LfD}{Learning from Demonstration}
\newacronym{vae}{VAE}{Variational Auto-Encoder}
\newacronym{kl}{KL}{Kullback–Leibler}
\newacronym{ssl}{SSL}{Self-Supervised Loss}
\newacronym{rmse}{RMSE}{Root Mean Squared Error}
\newacronym{pca}{PCA}{Principal Component Analysis}
\newacronym{dof}{DoF}{Degrees of Freedom}
\newacronym{tsne}{t-SNE}{t-distributed Stochastic Neighbor Embedding}
\newacronym{mae}{MAE}{masked autoencoder}
\newacronym{byol}{BYOL}{Bootstrap Your Own Latent}
\newacronym{relu}{ReLU}{Rectified Linear Unit}
\newacronym{seq2seq}{Seq2Seq}{Sequence-to-Sequence}
\newacronym{lstm}{LSTM}{Long Short-Term Memory}

\providecommand{\forcez}{\mathnormal{F_z}} 
\providecommand{\forcex}{\mathnormal{F_x}} 
\providecommand{\threshold}{\mathnormal{F_{x}^{threshold}}} 
\providecommand{\targetforce}{\mathnormal{F_{z}^{target}}} 
\providecommand{\admittanceconstant}{\mathnormal{\alpha}} 

\providecommand{\actionxref}{\mathnormal{u_{x}^{const}}} 
\providecommand{\actionup}{\mathnormal{u_{z}^{up}}} 
\providecommand{\actiondown}{\mathnormal{u_{z}^{down}}} 

\providecommand{\desirednextstate}{\mathnormal{x^*_{t:t+\nexttimesteps}}} 
\providecommand{\predictednextstate}{\mathnormal{\hat{x}_{t:t+\nexttimesteps}}} 
\providecommand{\truenextstate}{\mathnormal{x^{obs}_{t:t+\nexttimesteps}}} 
\providecommand{\nextstate}{\mathnormal{x_{t:t+\nexttimesteps}}} 
\providecommand{\paststate}{\mathnormal{x_{t-\prevtimesteps:t-1}}} 
\providecommand{\desirednextaction}{\mathnormal{u^*_{t:t+\nexttimesteps}}} 
\providecommand{\predictednextaction}{\mathnormal{\hat{u}_{t:t+\nexttimesteps}}} 
\providecommand{\truenextaction}{\mathnormal{u^{prim}_{t:t+\nexttimesteps}}} 
\providecommand{\nextaction}{\mathnormal{u_{t:t+\nexttimesteps}}} 
\providecommand{\stateactionconstant}{\mathnormal{\beta}} 

\providecommand{\cellstate}{\mathnormal{c}} 
\providecommand{\hiddenstate}{\mathnormal{h}} 

\providecommand{\lstmencoder}{\mathnormal{\theta}} 
\providecommand{\lstmdecoder}{\mathnormal{\phi}} 
\providecommand{\lstmencoderFC}{\mathnormal{\omega_{enc}}} 
\providecommand{\lstmdecoderFC}{\mathnormal{\omega_{dec}}} 

\providecommand{\nexttimesteps}{\mathnormal{T_n}} 
\providecommand{\prevtimesteps}{\mathnormal{T_p}} 

\providecommand{\policy}{\mathnormal{\pi}} 
\providecommand{\basepolicy}{\mathnormal{\pi_{b}}} 
\providecommand{\finetunedpolicy}{\mathnormal{\pi_{task}}} 

\providecommand{\demodata}{\mathnormal{D_{task}}} 
\providecommand{\primitivedata}{\mathnormal{D_{prim}}} 

\providecommand{\traj}{\mathnormal{\tau}} 
\providecommand{\primdatanum}{\mathnormal{N}} 
\providecommand{\trajlength}{\mathnormal{T}} 

\usepackage{amsmath}
\usepackage{amsfonts}

\usepackage{siunitx}

\usepackage{graphicx}
\usepackage{multirow}
\usepackage{hhline}
\usepackage{adjustbox}
\usepackage{tabularray}

\usepackage{subcaption}

\usepackage{makecell}

\usepackage{bm}

\usepackage[inline]{enumitem}
\usepackage{doi}

\usepackage{xcolor}

\usepackage{cite}

\newcommand{\paraDraft}[1]{\ifdefined\draft\subsubsection*{\color{blue}\textbf{#1}}\fi}

\providecommand{\eg}{\textit{e.g.,}~} %
\providecommand{\ie}{\textit{i.e.,}~} %
\providecommand{\sectionname}{Section}
\providecommand*{\sref}[1]{\sectionname~\ref{s:#1}}            
\providecommand{\tablename}{Table}
\providecommand*{\tref}[1]{\tablename~\ref{t:#1}}   
\providecommand{\figurename}{Fig.}
\providecommand*{\fref}[1]{\figurename~\ref{f:#1}}  
\providecommand*{\fref}[1]{\figurename~\ref{f:#1}}  
\providecommand{\equationname}{Eq.}
\providecommand*{\eref}[1]{\equationname~(\ref{e:#1})}            

\title{\LARGE \bf
Few-shot transfer of tool-use skills using human demonstrations with proximity and tactile sensing
}

\author{Marina Y. Aoyama$^{1}$, Sethu Vijayakumar$^{2}$, and Tetsuya Narita$^{3}$%
\thanks{Manuscript received: January, 21, 2025; Revised April, 21, 2025; Accepted June, 6, 2025.}
\thanks{
This paper was recommended for publication by Editor Ashis Banerjee upon evaluation of the Associate Editor and Reviewers' comments. 
} 
\thanks{$^{1}$School of Informatics, The University of Edinburgh, U.K.
{\tt\footnotesize m.aoyama@sms.ed.ac.uk}
}
\thanks{$^{2}$School of Informatics, The University of Edinburgh, U.K. {\tt\footnotesize sethu.vijayakumar@ed.ac.uk}}
\thanks{$^{3}$Sony Group Corporation, Japan. {\tt\footnotesize Tetsuya.A.Narita@sony.com}}
\thanks{
The first author conducted this research during an internship at Sony Group Corporation, Japan. 
The second author is partially funded by the Moonshot R\&D Program (Grant No. JPMJMS2031). 
We gratefully acknowledge Takahisa Ueno for his support with the setup of simulation and hardware. } 
\thanks{Digital Object Identifier (DOI): ~\href{https://doi.org/10.1109/LRA.2025.3583608}{\color{blue}10.1109/LRA.2025.3583608}.}
}
\begin{document}

\maketitle

\setlength{\textfloatsep}{0pt}
\setlength{\floatsep}{0pt}
\setlength{\intextsep}{0pt}

\titlespacing*{\subsection}{0pt}{5pt}{5pt}

\begin{abstract}
Tools extend the manipulation abilities of robots, much like they do for humans. 
Despite human expertise in tool manipulation, teaching robots these skills faces challenges. 
The complexity arises from the interplay of two simultaneous points of contact: one between the robot and the tool, and another between the tool and the environment.  
Tactile and proximity sensors play a crucial role in identifying these complex contacts. 
However, learning tool manipulation using these sensors remains challenging due to limited real-world data and the large sim-to-real gap. 
To address this, we propose a few-shot tool-use skill transfer framework using multimodal sensing. 
The framework involves pre-training the base policy to capture contact states common in tool-use skills in simulation and fine-tuning it with human demonstrations collected in the real-world target domain to bridge the domain gap.
We validate that this framework enables teaching surface-following tasks using tools with diverse physical and geometric properties with a small number of demonstrations on the Franka Emika robot arm. 
Our analysis suggests that the robot acquires new tool-use skills by transferring the ability to recognise tool-environment contact relationships from pre-trained to fine-tuned policies. 
Additionally, combining proximity and tactile sensors enhances the identification of contact states and environmental geometry. 
See our videos at~\href{https://sony.github.io/tool-use-few-shot-transfer/}{\color{blue}https://sony.github.io/tool-use-few-shot-transfer/}. 
\vspace{-0.3cm}
\end{abstract}


\section{INTRODUCTION}
\label{s:introduction}

\paraDraft{Importance of tool use}
\IEEEPARstart{T}{ools} extend the ability of robots to interact with the environment and manipulate objects~\cite{Qin_Brawer_Scassellati_2023}. 
For instance, mastering tool-use skills is crucial to perform complex tasks such as cutting vegetables with a knife, cleaning surfaces with a sponge, or picking up small screws with tweezers. 
This study focuses on manipulating grasped tools in a similar way to humans, rather than attaching tools to the end-effector as considered in most prior work~\cite{Aoyama_Moura_Saito_Vijayakumar_ICRA2023_wiping_pretraining, saito2021_ral_namiko_bestpaper, 
sundaresan2022_corl_skewering, Wi_Zeng_Florence_Fazeli_CoRL2022_virdo_extrinsic_tooluse_spatula, Yamane_Saigusa_Sakaino_Tsuji_RAL2023_graspedwriting_bilateral}. 
The ability to manipulate grasped tools enables robots to use tools designed for humans and seamlessly swap tools for different tasks. 

One of the challenges in acquiring tool-use skills is the complex contact relationships among the robot, the tool, and the environment. 
When a robot manipulates grasped tools, there are two locations of contact: the grasped point and the tool tip, as illustrated in~\fref{toolmanipulation_setup}. 
Intrinsic contact sensing involves direct contact between the tactile sensors and the tool. 
Extrinsic contact sensing~\cite{Ma_Dong_Rodriguez_ICRA2021_ExtrinsicContact} involves indirect sensing of the contact between the tool and the environment, with the tactile sensors being only in contact with the tool. 
The interplay between these two points of contact introduces complexity in sensing. 
Tactile sensing has proven useful in contact-rich tasks~\cite{Kim_Jha_Romeres_Patre_Rodriguez_icra2023_extrinsic_estimateandcontrol, guzey2023_arxiv_tactile_pretrain}, including tool-use scenarios~\cite{Shirai_Jha_Raghunathan_Hong_ICRA2023_tactiletool_extrinsic, Bronars_Kim_Patre_Rodriguez_icra2024_TEXterity, saito2021_ral_namiko_bestpaper}, while proximity sensing provides information about local geometry and the pose of the robot's fingertips relative to the environment~\cite{Navarro_MühlbacherKarrer_Alagi_Zangl_Koyama_Hein_Duriez_Smith_TRO2022_proximitysurvey, Sasaki_Koyama_Ming_Shimojo_Plateaux_Choley_IROS2018}. 
Despite these capabilities, there are currently no sensors capable of directly measuring extrinsic contact. 
Thus, the challenge lies in utilising multiple indirect sensor data sources to accurately identify these interactions across various tools and environments. 
To address this, we aim to combine tactile and proximity sensing to implicitly capture both intrinsic and extrinsic contacts, without requiring explicit disentanglement. 

\begin{figure}[t]
\centering
\includegraphics[keepaspectratio, scale=0.46]{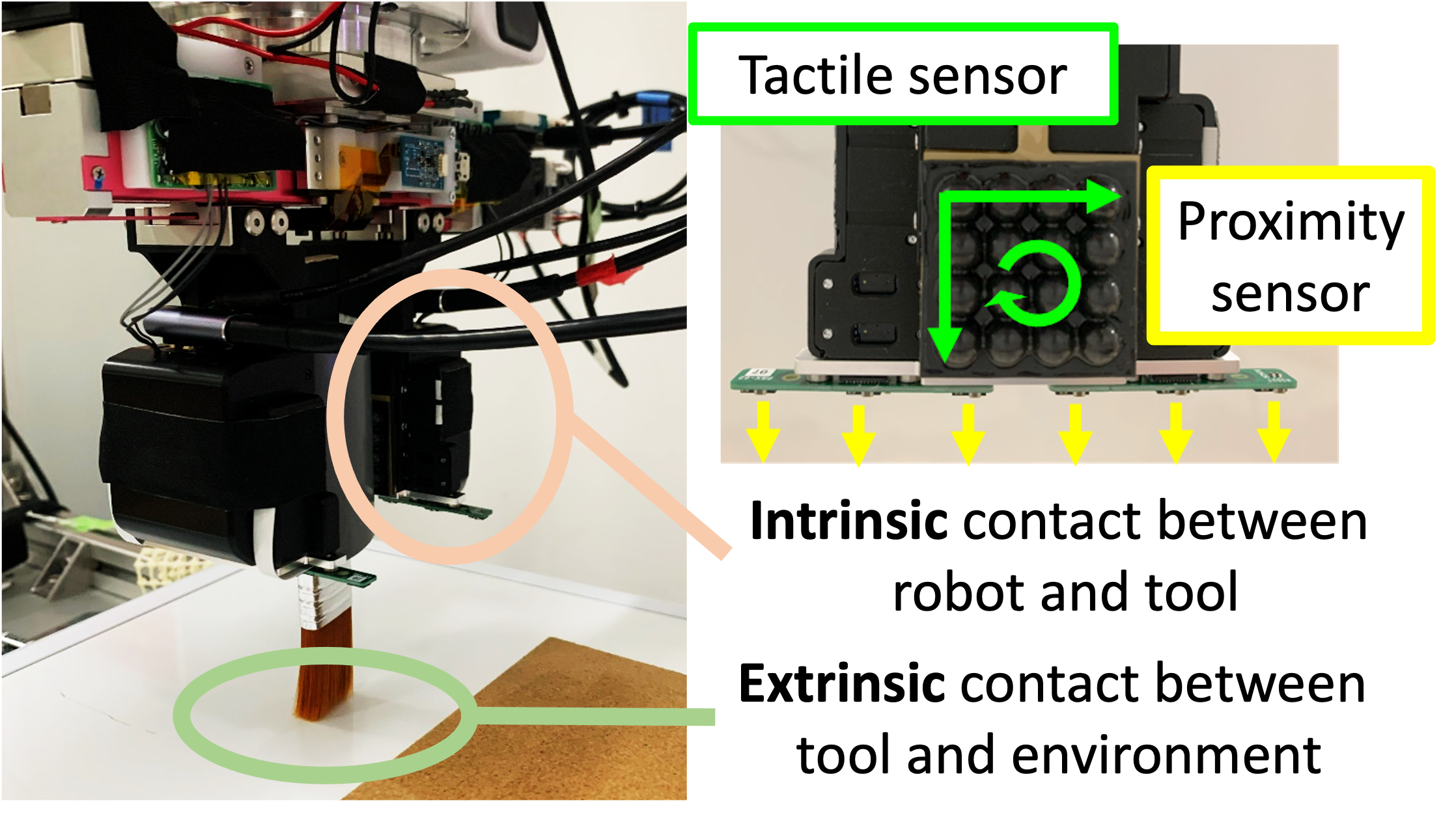}
\caption{
Tool manipulation using tactile and proximity sensors, involving two points of contact. 
}
\label{f:toolmanipulation_setup}
\end{figure}



To identify intrinsic and extrinsic contacts and manipulate deformable tools where detailed models or simulations of complex environments are unavailable, \gls{lfd} provides a viable approach for robots to acquire skills or behaviours by imitating human demonstrations\cite{Billard2008_lfdsurvey_efficient_intuitive}. 
However, the number of demonstrations is often limited due to the significant time and effort required from humans. 
It is, therefore, impractical to learn tool-use skills solely from demonstrations, given the wide-ranging variations in the physical and geometric properties of tools and environments. 
Alternatively, while learning in simulation offers the advantage of collecting large dataset at a low cost of time and human effort, it suffers from the sim-to-real gap. 


To enable robots to acquire various tool-use skills across different tools and task objectives, we propose a multimodal few-shot tool-use skill transfer framework. 
This approach involves two main phases: first, pre-training the base policy with primitive motions common in tool-use tasks in simulated environments; second, fine-tuning the policy with human demonstrations collected in the target domain. 
The key idea is to leverage inexpensive simulated data, which may differ from the target tool properties, environments, or tasks, while bridging this gap between the simulated pre-trained domain and the real target domain with human demonstrations. 
While the proposed framework is a general multimodal approach, we focus on the combined use of tactile and proximity sensors for tool-use skill acquisition.

\paraDraft{Experimental results}
We validate the proposed few-shot tool-use skill transfer framework on a surface-following task using tools with varying physical and geometric properties, such as brushes and sponges. 
This task serves as an exemplar of contact-rich tool manipulation tasks, where desired contact and motion depend on tool and task properties, and success requires fine control of contact.
Our results demonstrate that fine-tuning the policy pre-trained in simulation with human demonstrations enables the robot to manipulate new tools, adapt to different environments, and perform new tasks, ourperforming baseline approaches without pre-training or fine-tuning. 
Moreover, our analysis suggests the critical role of both proximity and tactile sensors in identifying contact force and the environment's geometry, particularly in the presence of sensor noise and the significant sim-to-real gap. 
Furthermore, a \gls{pca} of the model's latent space supports our hypothesis that the pre-trained policy transfers its ability to recognise tool-environment contact relationships across various tools.


\paraDraft{Contributions}
In summary, the contributions of this work are:
\begin{itemize}
    \item We tackle the challenging problem of learning tool-use skills that require delicate contact control, involving both intrinsic contact between the robot's fingertips and the tool and extrinsic contact between the tool and the environment.
    \item We present a data-efficient framework that utilises proximity and tactile sensing for pre-training a base policy in simulation and fine-tuning it in the real world to adapt to tools with diverse geometric and physical properties.
    \item We demonstrate that the proposed few-shot tool-use skill transfer framework achieves motion and contact forces close to those demonstrated by a human, successfully performing tool-use tasks on a physical robotic setup without fixing the tool to the end-effector.
\end{itemize}

\section{RELATED WORK}
\label{s:related_work}
\subsection{Tool manipulation}
\label{s:tool_use}
\paraDraft{Robot tool manipulation}
Tool-use skills are essential in many robot manipulation applications such as wiping~\cite{Aoyama_Moura_Saito_Vijayakumar_ICRA2023_wiping_pretraining}, transferring~\cite{saito2021_ral_namiko_bestpaper}, scraping~\cite{Merwe_Berenson_Fazeli_CoRL2022_extrinsic_tooluse_spatula_pretraining, Wi_Zeng_Florence_Fazeli_CoRL2022_virdo_extrinsic_tooluse_spatula}, and skewering~\cite{sundaresan2022_corl_skewering}.
However, most existing work simplifies these tasks by fixing the tool to the robot's end-effector, thereby avoiding the complex interactions between intrinsic and extrinsic contacts~\cite{Aoyama_Moura_Saito_Vijayakumar_ICRA2023_wiping_pretraining, saito2021_ral_namiko_bestpaper, sundaresan2022_corl_skewering, Wi_Zeng_Florence_Fazeli_CoRL2022_virdo_extrinsic_tooluse_spatula}. 
For robots to seamlessly swap and use various tools, they must be capable of grasping and manipulating tools. 

Several recent works successfully estimate and control extrinsic contacts between a tool~\cite{Shirai_Jha_Raghunathan_Hong_ICRA2023_tactiletool_extrinsic} or object~\cite{Kim_Jha_Romeres_Patre_Rodriguez_icra2023_extrinsic_estimateandcontrol, Bronars_Kim_Patre_Rodriguez_icra2024_TEXterity} and the environment. 
However, these approaches~\cite{Shirai_Jha_Raghunathan_Hong_ICRA2023_tactiletool_extrinsic, Kim_Jha_Romeres_Patre_Rodriguez_icra2023_extrinsic_estimateandcontrol, Bronars_Kim_Patre_Rodriguez_icra2024_TEXterity} are limited to the manipulation of rigid objects making contact with a flat surface due to the need to model the mechanics of contact. 
In contrast, our approach builds on research that learns the contact states between the tool and the environment from interaction data~\cite{Merwe_Berenson_Fazeli_CoRL2022_extrinsic_tooluse_spatula_pretraining, Hanwen_Zeyu_Wenyu_Haoyong_Yao_Yan_RAL2023_deformable_tool}, enabling us to manipulate deformable tools and contacts with non-flat surfaces that are difficult to model. 
However, these prior works~\cite{Merwe_Berenson_Fazeli_CoRL2022_extrinsic_tooluse_spatula_pretraining,Hanwen_Zeyu_Wenyu_Haoyong_Yao_Yan_RAL2023_deformable_tool} are limited to the manipulation of the tools used during training. 
No prior approaches have demonstrated the ability to perform tool manipulation with unattached, deformable tools across multiple tool types.

\subsection{Learning from Demonstration}
\label{s:lfd_contact}
\paraDraft{LfD for tool manipulation}
\gls{lfd} has been successfully used for acquiring tool manipulation skills~\cite{Aoyama_Moura_Saito_Vijayakumar_ICRA2023_wiping_pretraining, saito2021_ral_namiko_bestpaper, sundaresan2022_corl_skewering, Wi_Zeng_Florence_Fazeli_CoRL2022_virdo_extrinsic_tooluse_spatula, Yamane_Saigusa_Sakaino_Tsuji_RAL2023_graspedwriting_bilateral}. 
However, due to the time and human effort required for collecting demonstrations, the available demonstration data tends to be limited in both quantity and tool variation. 
One promising approach to enhance the robustness and generalisability of learnt skills, especially with limited demonstration data, is pre-training perceptual models~\cite{guzey2023_arxiv_tactile_pretrain, Aoyama_Moura_Saito_Vijayakumar_ICRA2023_wiping_pretraining} or state prediction models~\cite{Wi_Zeng_Florence_Fazeli_CoRL2022_virdo_extrinsic_tooluse_spatula, Lee2020_Trans_haptic_self-supervised}. 
However, pre-training using real-world play data collected through human teleoperation~\cite{guzey2023_arxiv_tactile_pretrain} or unsupervised interactions based on random and heuristic motions~\cite{Wi_Zeng_Florence_Fazeli_CoRL2022_virdo_extrinsic_tooluse_spatula, Lee2020_Trans_haptic_self-supervised} demands extensive human effort or raises the possibility of unsafe contacts. 
On the other hand, pre-training in simulation suffers from the sim-to-real gap, which refers to the discrepancy between simulated and real-world environments~\cite{Aoyama_Moura_Saito_Vijayakumar_ICRA2023_wiping_pretraining}. 
While there is a framework enabling zero-shot sim-to-real transfer~\cite{Wi_Merwe_Florence_Zeng_Fazeli_2023_calamari}, it only predicts contact patches and does not allow for precise control of contact force between tools and the environment. 

In our approach, we leverage learning the interaction model from inexpensive data collected safely in various simulated environments, with the capability to fine-tune this model using demonstration data from the real target domain. 
While there exists a framework for composing tool-use skills in multi-task and multi-domain learning contexts~\cite{Wang_Zhao_Du_Adelson_Tedrake_arXiv2024_poco}, it struggles in tasks requiring delicate contact control due to the domain gap between simulation or video and reality. 

\subsection{Tactile and proximity sensing}
\label{s:tactile_proximity}
\paraDraft{Tactile and proximity sensing}
Tactile sensors determine contact forces, contact events, local geometries, and material properties when they are in contact with the object or the environment~\cite{Li_Kroemer_Su_Veiga_Filipe_Kaboli_Ritter_TRO2020_tactilesurvey}, while proximity sensors identify local geometries and positions relative to the environment without contact~\cite{Navarro_MühlbacherKarrer_Alagi_Zangl_Koyama_Hein_Duriez_Smith_TRO2022_proximitysurvey}. 
Most existing methods use proximity sensors for the pre-contact phase~\cite{Sasaki_Koyama_Ming_Shimojo_Plateaux_Choley_IROS2018} and tactile sensors for the post-contact phase~\cite{Li_Zhang_Zhu_Wang_Lee_Xu_Adelson_FeiFei_Gao_Wu_corl2022_seefeelhear}. 
In contrast, we propose a configuration of proximity sensors that enables environmental observation while grasping and manipulating a tool, complemented by tactile sensors at the fingertips.
We hypothesise that this configuration provides additional information about the geometry (\eg shape and distance) of the environment around the tool-environment contact area. 
While multimodal sensing, such as vision and tactile sensing~\cite{Lee2020_Trans_haptic_self-supervised}, has shown improved manipulation performance, the simultaneous use of tactile and proximity sensing for tool manipulation involving intrinsic and extrinsic contacts remains underexplored.

\section{PROBLEM FORMULATION}
\label{s:problem}
We address the problem of learning tool-use skills involving two locations of contact: one between the robot and the tool, and the other between the tool and the environment, as illustrated in~\fref{toolmanipulation_setup}.
We consider a scenario where the physical and geometric properties of the tool and surface are unknown a priori. 
Therefore, the robot needs to identify them from multimodal sensory information on the fly and adapt the motions accordingly. 
Additionally, as the desirable contact force between the tool and the environment depends on the task, the robot needs to learn this force from demonstrations of the target task.
More specifically, the goal is to enable robots to learn a policy $\policy$ to predict the next desired actions $\predictednextaction$ for the next $\nexttimesteps$ time steps, similar to a human demonstrator, given tactile sensor observations $\mathnormal{x^{tac}_{t-\prevtimesteps:t-1}}$ and proximity sensor observations $\mathnormal{x^{prox}_{t-\prevtimesteps:t-1}}$ from the past $\prevtimesteps$ time steps.
In this study, we focus on surface-following tasks using grasped tools. 
These tasks require fine control of the contact force between the tool and the environment, considering the properties of the tool and the geometries of the surfaces. 
Examples include painting with a brush and sweeping with a broom.

\section{METHOD}
\label{s:method}


\begin{figure}[t]
\centering
\includegraphics[keepaspectratio, scale=0.5]{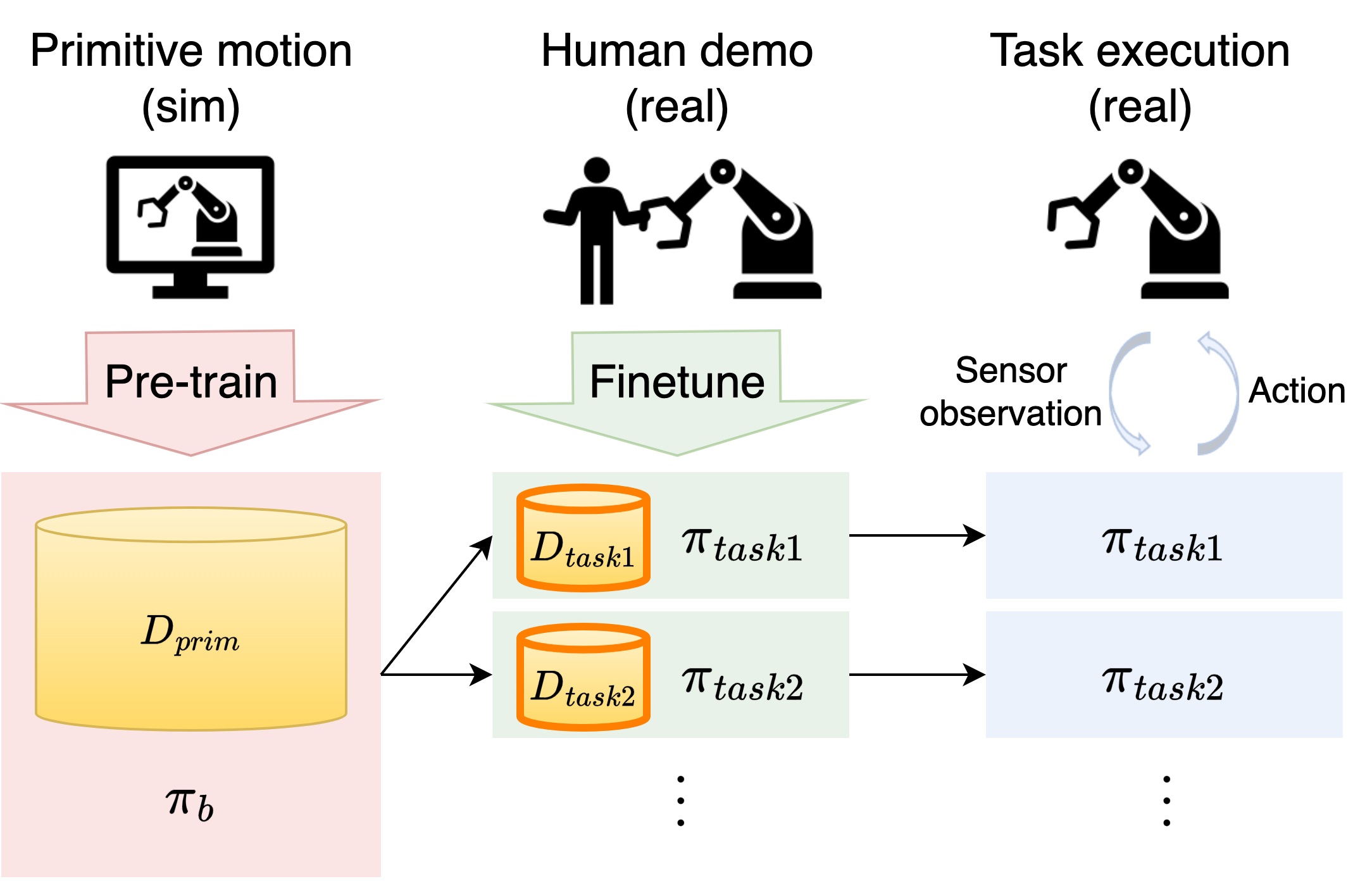}
\caption{
Few-shot tool-use skill transfer framework. 
\\
}
\label{f:policy_transfer_framework}
\vspace{0.2cm}
\end{figure}


\paraDraft{Overall method}
To enable the learning of new tool-use skills from a limited number of demonstrations, we propose a few-shot tool-use transfer framework. 
This framework involves pre-training the base policy using primitive motions in simulation and then fine-tuning it with human demonstrations of downstream tasks using the target tool in the real world to bridge the gap. 

\subsection{Pre-training base policy}
\label{s:pretraining_method}

In the pre-training phase, we train the base policy $\basepolicy$ by leveraging a large amount of data $\primitivedata=\left \{ \traj_1, \cdots \traj_\primdatanum \right \}$, where $\traj = \{(x_0, u_0), (x_1, u_1), \ldots, (x_\trajlength, u_\trajlength)\}$ collected through primitive motions in simulation. 
Here, $\primdatanum$ denotes the number of trajectories, and $\trajlength$ represents the number of timesteps in each trajectory. 
By performing this set of primitive motions repeatedly in simple simulated environments with varying conditions, we expect the base policy $\basepolicy$ to recognise contact states between the tool and the environment from multi-modal sensor observations and to capture motions common in tool-use skills, in a manner transferable across different tasks and tools. 
For instance, in surface-following tasks, we hypothesise that contact states such as contact angle and tooltip contact with a wall, and motions such as tracking a desired contact force or lifting up when the tooltip collides with a wall, are common to all surface-following tasks regardless of the tool properties and task goals.
It is important to highlight that the proposed framework requires primitive data collection and pre-training only once, as illustrated in~\fref{policy_transfer_framework}. 
Every new task with a new tool, environment, or task goal shares this single base policy. 
For the surface-following task, we use pre-defined primitive motions to collect data in simulation, as explained in~\sref{primitive_datacollection}. 
However, we emphasise that our framework can utilise other low-cost data sources, such as interaction data gathered through teleoperation or autonomous exploration. 

\begin{figure}[t]
\centering
\includegraphics[keepaspectratio, scale=0.43]{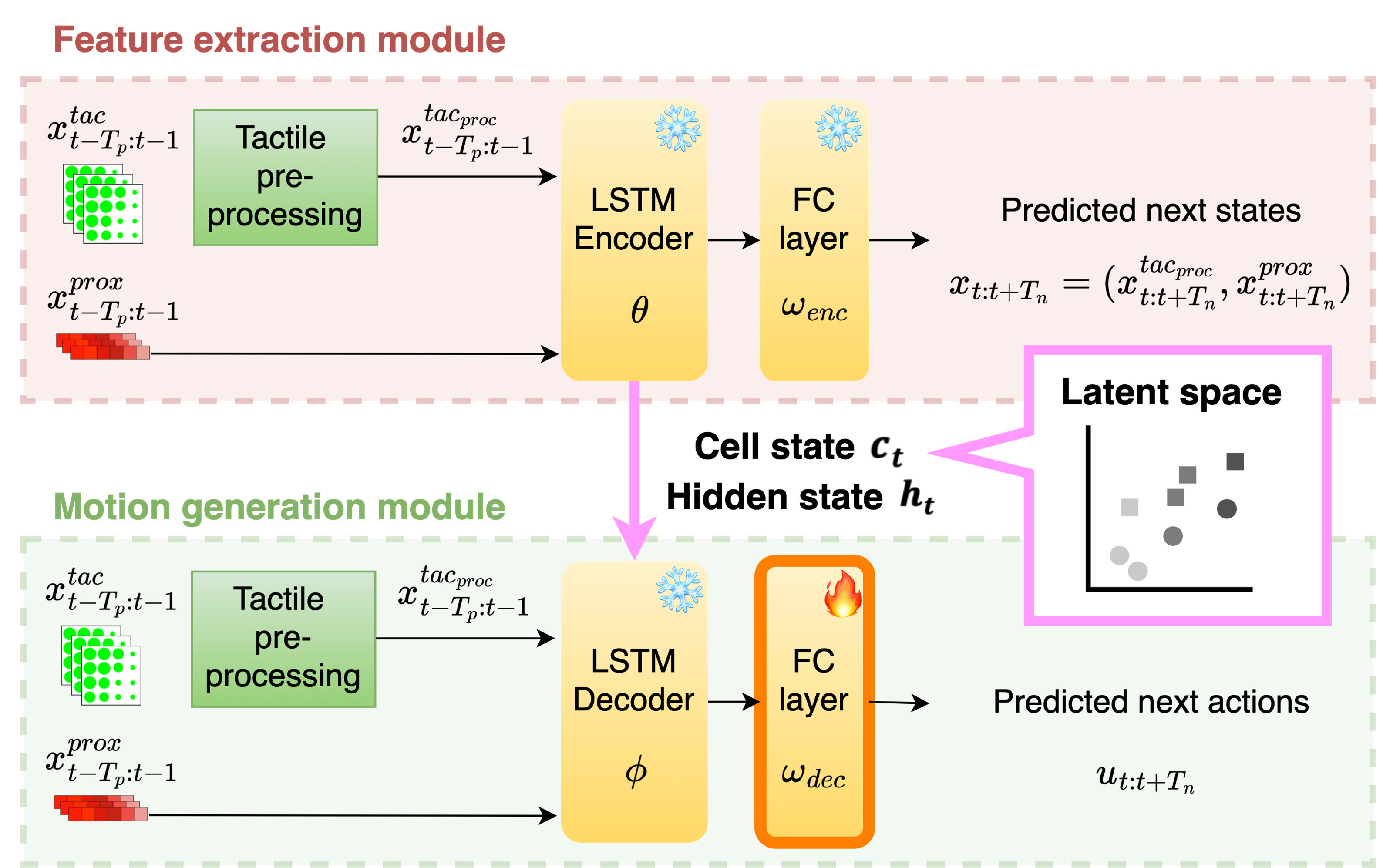}
\caption{
Seq2Seq architecture of the tool-use skill policy.  
The feature extraction module captures the contact relationship between the robot, tool, and environment by predicting next states. 
The motion generation module predicts the next desired actions to perform the task.
}
\label{f:seq2seq_architecture}
\vspace{0.1cm}
\end{figure}
For this, we adopt the \gls{seq2seq} model~\cite{Sutskever_Vinyals_Le_NeurIPS2014_seq2seq}, a widely used time-series prediction model in natural language processing~\cite{Cho_Merrienboer_GulcehreBahdanau_Bougares_Schwenk_Bengio_emnlp2014_seq2seq_nlp} and robot control~\cite{Kutsuzawa_Sakaino_Tsuji_RAL2018_seq2seq_robot}. The model employs an encoder-decoder structure using \gls{lstm}, as illustrated in \fref{seq2seq_architecture}.
The encoder serves as a feature extraction module. 
This module encodes multi-modal sensor observations $x$ from the past $\prevtimesteps$ timesteps into latent representations consisting of cell and hidden states by predicting the next $\nexttimesteps$ states $\predictednextstate$. 
In our case, with tactile and proximity sensors, the sensor observations $x$ include $x_{tac}^{t-T_p}$ and $x_{prox}^{t-T_p}$.
The decoder functions as the motion generation module, predicting the desired actions $\truenextaction$ in specified in primitive motions for the next $T_n$ timesteps. 
We condition this motion generation on the encoded contact states, which are passed from the encoder's cell and hidden states to the decoder.
The encoder-decoder structure offers key advantages for our framework. 
The encoder extracts latent representations of contact dynamics, decoupled from motion generation, enabling the reuse of the feature extraction module across tasks. 
Meanwhile, we can fine-tune the decoder for task-specific motion generation, allowing data-efficient adaptation to new tasks while retaining the generalisable features from the encoder. 
To achieve this, we train the model on primitive data by minimising the pre-training loss
\begin{align}
&\hat{\lstmencoder}, \hat{\lstmdecoder}, \hat{\lstmencoderFC}, \hat{\lstmdecoderFC} = 
\mathop{\arg \min}\limits_{\lstmencoder, \lstmdecoder, \lstmencoderFC, \lstmdecoderFC} L\left ( \nextstate, \nextaction \right ) \label{e:pretrain_loss} \\
&= E_{MSE}\left ( \predictednextstate, \truenextstate \right ) + \stateactionconstant E_{MSE}\left ( \predictednextaction, \truenextaction \right ) \nonumber
\end{align}
where $\lstmencoder$ and $\lstmencoderFC$ denote the parameters of the LSTM layer and fully connected layer in the encoder, respectively, and $\lstmdecoder$ and $\lstmdecoderFC$ represent the parameters of the LSTM layer and fully connected layer in the decoder of the seq2seq model, respectively, as shown in~\fref{seq2seq_architecture}. 
The first term of the loss function is the state prediction loss, which computes the Mean Squared Error $E_{MSE}$ between the observed states $\truenextstate$ and the predicted states $\predictednextstate$ for the next $\nexttimesteps$ time steps to train the feature extraction module. 
The second term of the loss function is the action prediction loss, which computes the Mean Squared Error $E_{MSE}$ between the actions $\truenextaction$ observed during the execution of primitive motions and the predicted actions $\predictednextaction$ for the next $\nexttimesteps$ time steps to train the motion generation module. 
$\stateactionconstant$ is the weight coefficient that balances the two loss terms.
Although we chose Seq2Seq for its data efficiency and fast execution, our framework potentially extend to other encoder-decoder models, such as transformer-based approaches~\cite{zhao2023learning, chi2023diffusion}.

\subsection{Few-shot fine-tuning using demonstration}
\label{s:finetuning}
\paraDraft{Finetuning base policy}
Subsequently, we fine-tune the pre-trained base policy $\basepolicy$ to obtain the fine-tuned policy $\finetunedpolicy$ using a small amount of demonstration data $\demodata$ specific to each new downstream task. 
We acknowledge the differences in tool properties, environments, percepts, and task goals between the pre-trained domain and the target downstream task domain. 
We expect that fine-tuning using demonstration data collected in the target domain (\ie using the target tool to achieve the downstream task in the target environment) bridges this domain gap and adapt the policy to manipulate new tools, adapt to new environments, adapt from simulated to real environments, and achieve new task goals.

To prevent over-fitting when fine-tuning the policy with a very small amount of demonstration data in the target domain, we minimize the number of layers to fine-tune.
Specifically, we finetune only the fully connected (FC) layer of the decoder while freezing the weights of all other layers in both the encoder and decoder. 
Therefore, while the primary role of the decoder during pre-training is motion generation, we expect the updated weights of $\lstmdecoderFC$ to adapt not only to differences in task goals and motions but also to domain differences, such as tool properties and other gaps between the pre-trained and target task domains. 
We achieve this adaptation by incorporating both state prediction and action prediction losses, as shown in LfD loss 
\begin{align}
\hat{\lstmdecoderFC} &= \arg\min_{\lstmdecoderFC} L\left ( \nextstate, \nextaction \right ) \label{e:lfd_loss} \\
&\hspace{-2em} \quad = E_{MSE}\left ( \predictednextstate, \desirednextstate \right ) + \stateactionconstant E_{MSE}\left ( \predictednextaction, \desirednextaction \right ) \nonumber. 
\end{align}

\subsection{Task execution}
\label{s:taskexecution}
\paraDraft{Task execution using learnt policy}
At test time, the robot predicts the next desired actions $\predictednextaction$ based on past sensor observations $\paststate$ using the fine-tuned policy $\finetunedpolicy$ to perform the downstream task. 
At each timestep, the robot predicts actions for the next $\nexttimesteps$ time steps to facilitate smoother trajectories and capture the temporal dependencies of its actions, while executing only the action predicted for the immediate next timestep $t$. 



\section{EXPERIMENTAL SETUP}
\label{s:experiment_setup}
\renewcommand\thesubfigure{\alph{subfigure}}
\begin{figure*}[t]
\vspace{0.1cm}
\centering
\includegraphics[keepaspectratio, scale=0.65]{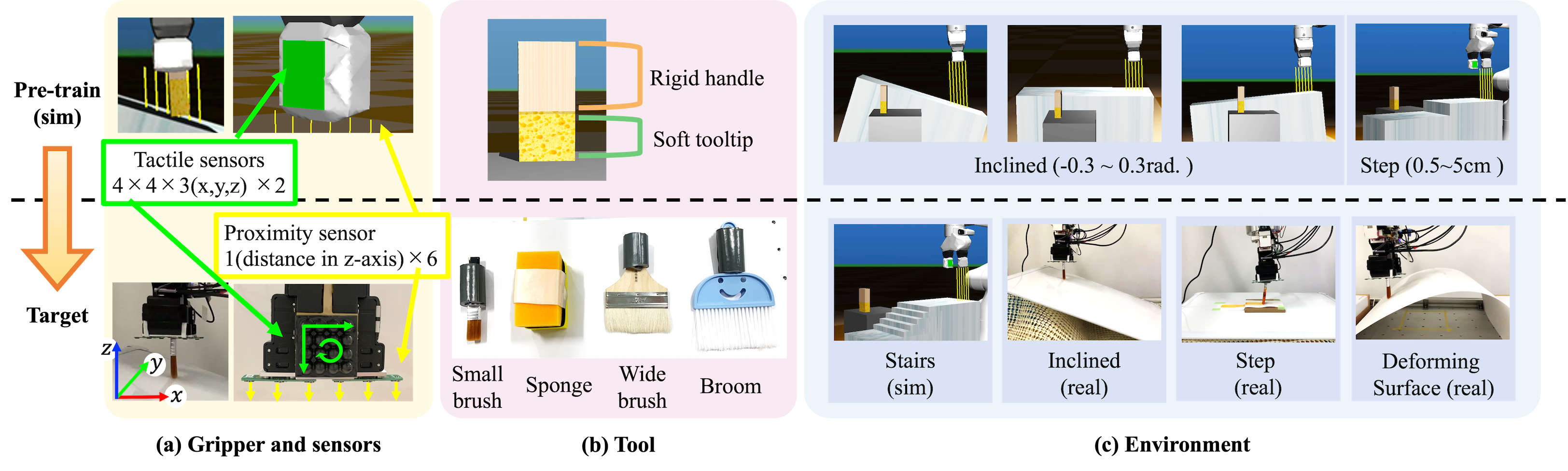}
\vspace{-0.2cm}
\caption{Experimental setup. } 
\label{f:experiment_setup}
\vspace{-0.6cm}
\end{figure*}

\subsection{Tool-use tasks}
\label{s:tooluse_tasks}
\paraDraft{Tools}

We conducted our experimental evaluation on surface-following tasks using tools with varying geometric (\eg tool length, contact area) and physical (\eg stiffness, friction) properties. 
In simulation, we designed a brush with a 3cm rigid handle and a soft tooltip (2×3×3 capsules, 2.5cm spacing, stiffness 30 using MuJoCo’s composite API) for pre-training. 
We then validated our proposed framework in real environments using four different tools: a small brush, a wide brush, a sponge, and a broom on a physical setup as depicted in~\fref{experiment_setup}. 
Each tool had a soft material around the handle to ensure a stable grasp. 
For the surfaces, we prepared two basic environments: “Inclined,” which is a flat surface inclined at $\psi=\left[ -0.3, 0.3 \right]$ radians, and “Step” which has a step with a height of $h=\left[ 0.5, 5.0 \right]$ cm in simulation. 
We tested in the 'Inclined' and 'Step' environments in the real world. 
Additionally, we tested more complex environments, including 'Stairs' with multiple consecutive steps in simulation and 'Deforming Surface' in the real world.

\vspace{-0.1cm}
\subsection{System design and robot setup}
\label{s:experimental_setup}
\paraDraft{Robot setup}
~\fref{experiment_setup} depicts the experimental setup and sensor configurations.
We utilise a 7 Degrees of Freedom (DoF) Franka Emika Panda robotic arm with a position controlled 2-finger parallel gripper attached to its end-effector for both simulation and real robot experiments. 
For the simulation setup, we utilise the Mujoco physics engine~\cite{Todorov_Erez_Tassa_iros2012_mujoco, Zakka_Tassa_menagerie2022github}, configured to replicate the physical robot setup. 
At the start of each trial, the robot grasps the tool handle near its centre, with a position shift sampled from \( \mathcal{N}(0, 1) \) cm, by closing the gripper until the normal force exceeds 5N, and then maintains a fixed gripper width throughout the trial. 
The base and fine-tuned policies output end-effector velocities in the x and z directions, as shown in~\eref{primitive_motion}, which we convert to end-effector positions and then map to joint configurations using inverse kinematics to control the robot. 
We provide demonstrations by kinaesthetically moving the robot arm in gravity compensation mode.


Inside each fingertip, we attach tactile sensors~\cite{Narita_Nagakari_Conus_Tsuboi_Nagasaka_ICRA2020_naritaslip} with 4 by 4 elastic hemispheres placed on 12 by 12 pressure sensor nodes. The tactile sensors estimate the 3-axis force at each hemisphere by measuring its distortion. 
In simulation, we use 3-axis force sensors arranged in a corresponding 4 by 4 configuration as an approximate match.
For both simulated and real tactile observations, we preprocess 3-axis forces from each fingertip, denoted as $x^{tac}$, into shear forces $x^{shear} \in \mathbb{R}^{2}$ and translational and rotational slip $x^{slip} \in \mathbb{R}^{3}$, following the theory of translational and rotational slip~\cite{Narita_Nagakari_Conus_Tsuboi_Nagasaka_ICRA2020_naritaslip}. This tactile representation, $x^{tac_{proc}} = \left( x^{shear}, x^{slip} \right)$, provides a compact and meaningful low-dimensional tactile feature that reduces the sim-to-real gap and enhances robustness against variations in raw sensor values caused by slight changes in grasp location or sensor noise.
Additionally, we place six Time Of Flight (ToF) proximity sensors on the right fingertip, oriented to look down at the environment. 
Each sensor measures the distance between itself and the closest object as a scalar value.

\subsection{Data collection}
\label{s:experimental_setup}
\subsubsection{Primitive data collection}
\label{s:primitive_datacollection}
\paraDraft{Primitive data collection}
First, we collect primitive data $\primitivedata$ for the surface-following task in two basic environments, "Inclined" and "Step", in simulation. 
For the surface-following task, we define primitive motions along the $x$-axis as $u_x = \actionxref$, and along the $z$-axis as
\begin{equation}
\begin{aligned}
u_z &= \begin{cases} 
\actionup & \text{if } \forcex > \threshold \text{ (hitting wall)}\\
\actiondown & \text{elif } \forcez < 0.0 \text{ (out of contact)}\\
\admittanceconstant(\targetforce - \forcez) & \text{otherwise (in contact)}  
\end{cases}
\end{aligned}
\label{e:primitive_motion}
\end{equation}
where we define the axis coordinates in the task frame, as illustrated in~\fref{experiment_setup}.
The robot moves its end-effector at a constant velocity $\actionxref$ in the horizontal direction. 
When it detects a collision with a wall, it moves the end-effector up by $\actionup$; 
if out of contact, it moves the end-effector down by $\actiondown$ to reestablish contact. 
Otherwise, it adjusts the velocity of the end-effector in the normal direction $u_z$ to achieve the specified target force $\targetforce$, without tool slippage, given the current normal force $\forcez$ and the scaling factor $\admittanceconstant$. 
In our experiment, we set parameters to $\actionxref =\SI[per-mode=symbol]{0.3}{\centi\meter\per\second}$, $\actionup = \SI[per-mode=symbol]{0.5}{\centi\meter\per\second}$, $\actiondown = \SI[per-mode=symbol]{-0.5}{\centi\meter\per\second}$, $\threshold = 0.5$ N, $\targetforce = 0.3$ N, and $\admittanceconstant = 0.1$.
We collected 11 trajectories in the “Inclined” environment and 10 trajectories in the “Step” environment.
For each trial, we record the end-effector positions, tactile data ($ 4 \times 4 \times 3 $ axes for 2 fingertips), and proximity data (6 dim.) at a frequency of 20 Hz over a duration of 10 seconds. 





\subsubsection{Demonstration data collection}
\label{s:demonstration_datacollection}
\paraDraft{Demonstration data collection}
We collect demonstration data $\demodata$ to fine-tune the base policy $\basepolicy$ for learning the target task motion. 
For the simulated experiments, we collect 3 demonstration trajectories each for
\begin{enumerate*}[label=(\arabic*)]
    \item the new target force where $\targetforce = 0.5N$ and 
    \item the stairs environment. 
\end{enumerate*} For the physical robot experiments, we collect 3 demonstration trajectories for each downstream task, using different tools. 
We record the end-effector positions, tactile data ($4\times4\times3$ axes for 2 fingertips), and proximity data (6 dim.) at a frequency of 20 Hz for 10 seconds.  


\subsection{Model training}
\label{s:model_training}
\paraDraft{Seq2seq parameters}
\label{s:model_parameters}
The seq2seq model consists of 1 LSTM layer and 1 fully connected layer in the encoder, and 1 LSTM layer and 1 fully connected layer in the decoder. 
We apply dropout~\cite{srivastava2014dropout} to the fully connected layers following the LSTM layers in both the encoder and decoder to prevent over-fitting during pre-training, with a dropout rate of 0.2. 
We set the dimensions of the cell state and hidden state to $\cellstate \in \mathbb{R}^{100}$ and $\hiddenstate \in \mathbb{R}^{100}$, respectively. 
We employ stochastic gradient descent~\cite{Ruder_arXiv2016_sgd} to update the weights of the neural network. 
We normalise all end-effector position data, shear force and slip data, and proximity data to the range $\left[-0.9, 0.9\right]$. 

\paraDraft{Seq2seq model training}
First, we pre-train the seq2seq model using primitive data as detailed in~\sref{pretraining_method}. 
We train the model for 2000 epochs at a learning rate of 0.001, utilising a loss function defined in~\eref{pretrain_loss} with $\stateactionconstant = 0.1$. 
It is important to note that the collection of primitive data and the pre-training of the base policy are only required once. 
This single pre-trained model serves as the base policy for all downstream tasks. 
After pre-training the base policy, we freeze the weights of all layers in the encoder and the LSTM layer in the decoder. 
For each task, we fine-tune the fully connected layer in the decoder on demonstration data for 300 epochs at a learning rate of 0.005 using the loss function~\eref{lfd_loss}.

\section{RESULTS AND DISCUSSION}
\label{s:result_and_discussion}
We evaluate the ability of the proposed few-shot tool-use skill transfer framework to adapt the robot's motion for new tools and target tasks with only a few demonstrations.  


We compared our proposed approach, denoted as 'Finetuned,' with a behaviour cloning baseline, "Demo only"~\cite{pomerleau1988alvinn}, which uses a policy trained solely on demonstration data without any pre-training in simulation, and an ablated baseline, "No FT," which uses the pre-trained policy without fine-tuning in the target domain. 
We also included two alternative methods used in settings with limited supervised data: Gaussian Processes (GP) with LSTM~\cite{al2017learning}, to incorporate temporal information for fair comparison, trained on demonstration data without pre-training; and 
Dataset Aggregation (DAgger)~\cite{ross2011reduction}, where we trained on pre-training data in the first iteration, and on aggregated data---including additional fine-tuning demonstrations as corrections---in the second.
In all approaches, we collected three demonstrations and empirically tuned hyperparameters to minimise the training loss.

For the “Inclined” environment in~\fref{experiment_setup}, we computed: 1) the \gls{rmse} between the ground truth and the followed inclination, and 2) the \gls{rmse} between the desired contact force and the applied force to evaluate how effectively the tool followed the surface geometry while applying the desired force. 
We collected demonstration data for test cases in addition to the training data to obtain the desired contact force for evaluation.
For surfaces with steps (i.e., "Step" and "Stairs" in~\fref{experiment_setup}), the \gls{rmse} metrics were unsuitable due to intentional tool lifting when hitting a step. 
Thus, we used wiped area (\%)---the contact area relative to the total surface area---as the evaluation metric. 
Since demonstrations aim to maximise the wiped area by quickly reestablishing contact after lifting the tool, this metric reflects both imitation accuracy and task completeness. 


\paraDraft{(C3) Transfer to downstream tasks} 
\subsection{Does the tool-use skill policy, fine-tuned with demonstrations, improve the performance of downstream tasks?} 
\paraDraft{Simulated experiments for new target force} 
First, we evaluated our approach in two test cases using the simulated environments "Inclined" and "Stairs". 
For the inclined surface, the task involved wiping an inclined surface with a desired force of 0.5N, which differed from the 0.3N used during pre-training.  
In the "Stairs" environment, the task was to wipe a surface with stairs instead of a single step seen in the pre-training phase.  

In both environments, "Finetuned" achieved the highest task performance compared to the baselines ("Demo only" and "No FT") as well as the alternative methods "GP" and "DAgger," as depicted in~\tref{sim_rmse_table}. 
~\fref{sim_example_traj_inclined} illustrates that "Demo only" failed to achieve the desired force due to insufficient demonstrations.  
In contrast, "Finetuned" successfully achieved the new desired contact force, different from the desired force used in pre-training. 
~\fref{sim_example_traj_stairs} shows the end-effector position in the z-direction as well as the contact force in the normal direction while wiping stairs using the fine-tuned policy. 
The red shading indicates the contact of the tool tip with the wall, and the blue indicates the flat region (\ie no contact of the tool tip with the wall).
We observed that the fine-tuned policy detects the contact of the tool tip with the wall, leading to successful lifting of the tool, and returns to the desired contact force (\ie 0.3N in this case) when reaching the flat region. 
We also evaluated the generalisability of the policy fine-tuned on a different tool instance with a longer handle (5cm vs. 3cm) and softer tip (stiffness 10 vs. 30) (FT (diff. tool) in~\tref{sim_rmse_table}); 
while this policy outperformed the baselines, the one fine-tuned on the target tool achieved the lowest error, highlighting the importance of task-specific fine-tuning.  

\begin{table}[t]
\vspace{0.2cm}
\centering
\scriptsize 
\setlength{\tabcolsep}{3pt} 
\begin{tabular}{c@{\hskip 3pt}c@{\hskip 3pt}c@{\hskip 3pt}c@{\hskip 3pt}c@{\hskip 3pt}c@{\hskip 3pt}c@{\hskip 3pt}c}
\hline
\textbf{Env.} & 
\makecell[c]{\textbf{Evaluation}\\\textbf{Criteria}} & 
\makecell[c]{\textbf{Demo}\\\textbf{only}} & 
\makecell[c]{\textbf{No}\\\textbf{FT}} & 
\makecell[c]{\textbf{FT}\\\textbf{(diff. tool)}} & 
\textbf{GP} & 
\textbf{DAgger} & 
\textbf{Finetuned} \\
\hline\hline
\multirow{3}{*}{\textbf{Inclined}} 
  & \makecell[c]{RMSE of\\slope (rad) $\bm{\downarrow}$} 
    & \makecell{0.073 \\ \scriptsize$\pm$ 0.002} 
    & \makecell{0.041 \\ \scriptsize$\pm$ 0.006} 
    & \makecell{0.038 \\ \scriptsize$\pm$ 0.003} 
    & \makecell{0.095 \\ \scriptsize$\pm$ 0.007} 
    & \makecell{0.072 \\ \scriptsize$\pm$ 0.005} 
    & \makecell{0.040 \\ \scriptsize$\pm$ 0.005} \\
  & \makecell[c]{RMSE of\\force (N) $\bm{\downarrow}$} 
    & \makecell{0.68 \\ \scriptsize$\pm$ 0.06} 
    & \makecell{0.22 \\ \scriptsize$\pm$ 0.01} 
    & \makecell{0.17 \\ \scriptsize$\pm$ 0.02} 
    & \makecell{0.61 \\ \scriptsize$\pm$ 0.05} 
    & \makecell{0.62 \\ \scriptsize$\pm$ 0.07} 
    & \makecell{\textbf{0.11} \\ \scriptsize$\pm$ 0.02} \\
\hline
\textbf{Stairs} 
  & \makecell[c]{Wiped\\area (\%) $\bm{\uparrow}$} 
    & \makecell{45.0 \\ \scriptsize$\pm$ 3.50} 
    & \makecell{49.7 \\ \scriptsize$\pm$ 2.31} 
    & \makecell{53.1 \\ \scriptsize$\pm$ 2.98} 
    & \makecell{43.3 \\ \scriptsize$\pm$ 4.01} 
    & \makecell{51.1 \\ \scriptsize$\pm$ 3.80} 
    & \makecell{\textbf{59.2} \\ \scriptsize$\pm$ 4.31} \\
\hline
\end{tabular}
\caption{
Comparison of \gls{rmse} when transferring the pre-trained tool-use skill to a new task (\ie with a new target force) and to a more complex environment (\ie stairs). 
}
\label{t:sim_rmse_table}
\vspace{0.1cm}
\end{table}

\begin{figure}[t]
\captionsetup[subfigure]{justification=centering}
    \begin{tabular}{cc}
      \begin{minipage}[b]{0.45\hsize}
        \centering
        \includegraphics[keepaspectratio, scale=0.26]{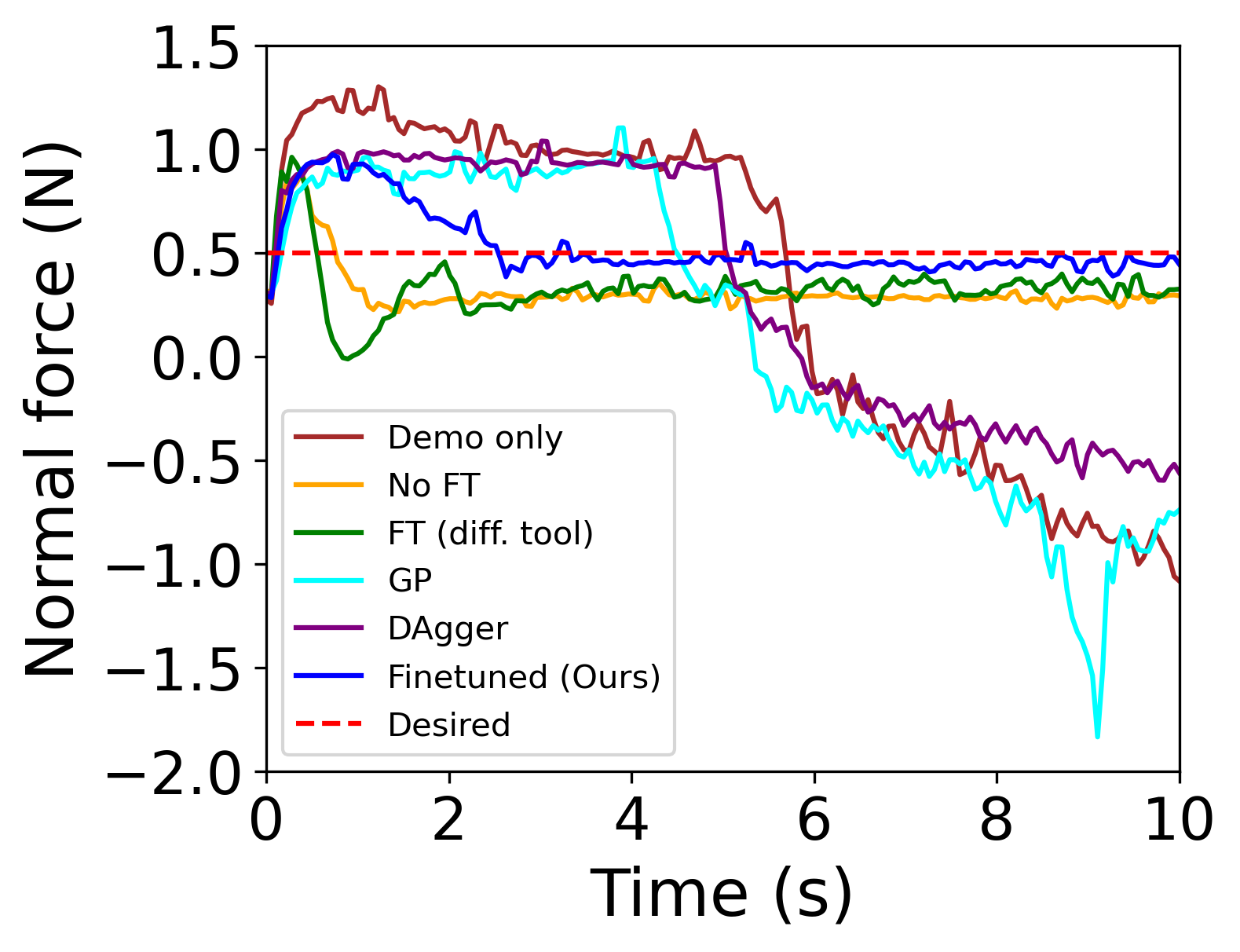}
        \vspace{-0.2cm} 
        \subcaption{Inclined}
        \label{f:sim_example_traj_inclined}
      \end{minipage} &
      \hspace{-0.7cm}
      \begin{minipage}[b]{0.45\hsize}
        \centering
        \includegraphics[keepaspectratio, scale=0.28]{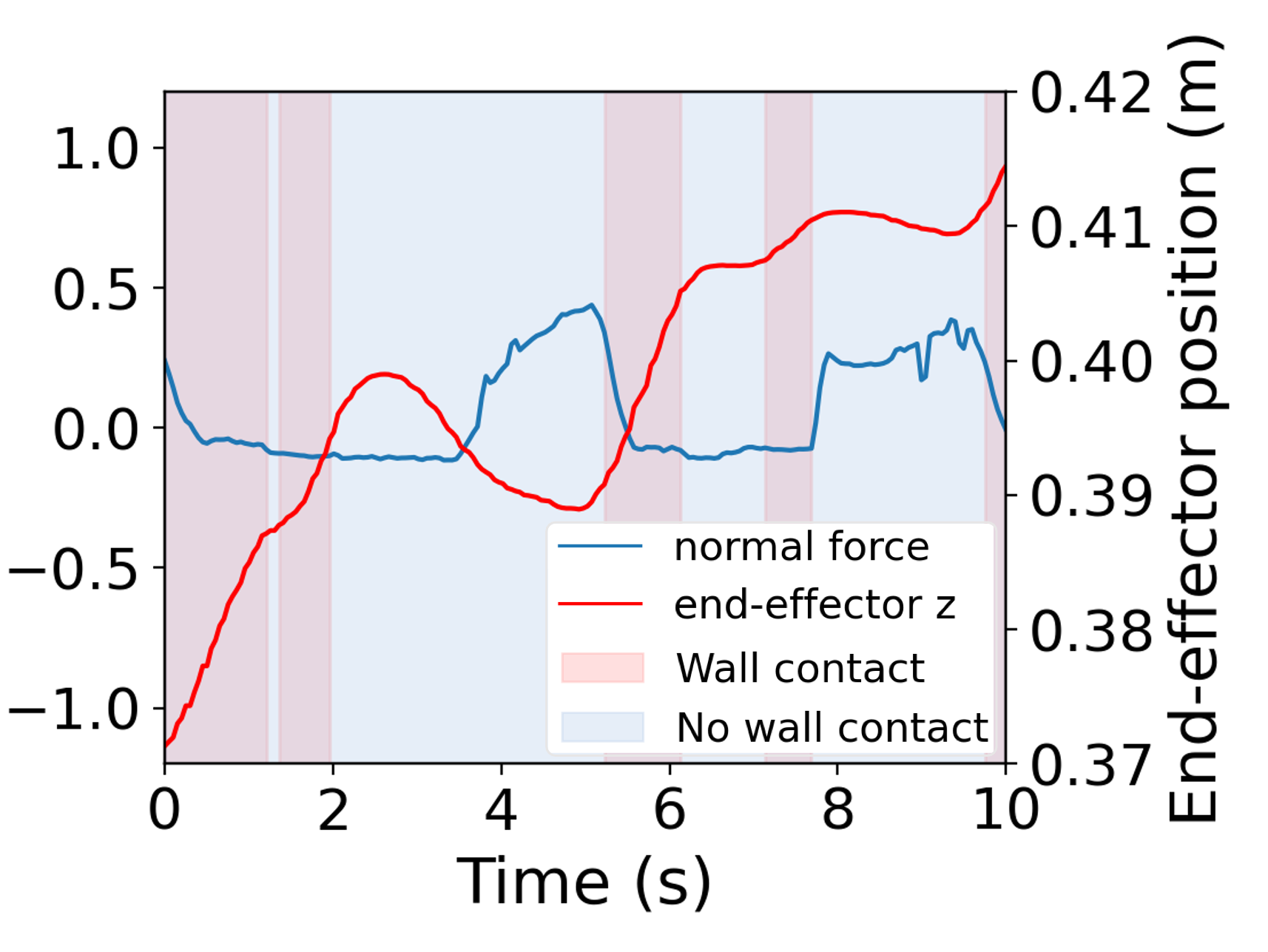}
        \vspace{-0.2cm} 
        \subcaption{Stairs}
        \label{f:sim_example_traj_stairs}
      \end{minipage}
    \end{tabular}
    \vspace{-0.3cm} 
     \caption{End-effector motion and contact force profile. 
     }
     \label{f:sim_example_trajectories}
\vspace{0.2cm}
\end{figure}

\paraDraft{Real experiments} 
Second, we evaluated our approach in two real-world environments: "Inclined" and "Step", using four different tools. 
In both environments, the robot needs to adapt its motions to real tools with various properties that are inherently different from those used in simulation, while also accounting for the sim2real gap in sensing. 
Additionally, the robot needs to achieve a new target contact force required for each task, as demonstrated by a human. 

\tref{real_evaluation_slope} and \tref{real_evaluation_force} display the \gls{rmse} of the inclination and the contact force, respectively, for four downstream tasks demonstrated by a human using different tools shown in~\fref{experiment_setup}. 
In all cases, "Finetuned" achieved a significant reduction in \gls{rmse} compared to the baselines. 
In real-world task performance, in the "small brush" painting task, "Finetuned" painted $93.5 \pm 1.68\%$ of the surface, outperforming "No FT" ($87.1 \pm 1.61\%$) and "Demo Only" ($72.7 \pm 4.75\%$). 
In the "broom" task, it swept $8.0 \pm 1.0$ dust pieces out of 10, compared to $2.7 \pm 1.53$ for "No FT" and $6.3 \pm 0.58$ for "Demo Only".
Furthermore, for the task of painting a surface with a step, "Finetuned" notably increased the wiped area, as evidenced by the results in~\tref{real_evaluation_wipedarea}. 

\paraDraft{Summary of real experiments} 
These results, where the proposed approach outperforms the "Demo only" and "No FT" baselines, demonstrate the ability of our few-shot tool-use skill transfer framework to manipulate various tools by: 
\begin{enumerate*}[label=(\arabic*)]
\item leveraging the base policy pre-trained in simulation, and
\item addressing the sim-to-real gap and differences in tool properties with human demonstrations collected in the target domain.
\end{enumerate*}

\begin{table}[t]
\centering 
\begin{subtable}{\linewidth}
\centering
\begin{tabular}{ccccccc}
\hline
\multirow{2}{*}{\textbf{Tool}} & \multicolumn{2}{c}{\textbf{Demo only}} & \multicolumn{2}{c}{\textbf{No FT}} & \multicolumn{2}{c}{\textbf{Fintuned}} \\
                               & x                  & $\sigma$                 & x                    & $\sigma$                   & x                 & $\sigma$                 \\ \hline\hline
Small brush                    & 0.41               & 0.16              & 0.25                 & 0.07                & \textbf{0.17}              & 0.02              \\
Sponge                         & 0.30               & 0.12              & 0.29                 & 0.01                & \textbf{0.17}              & 0.03              \\
Wide brush                     & 0.18               & 0.04              & 0.37                 & 0.12                & \textbf{0.10}              & 0.02              \\
Broom                          & 0.84               & 0.33              & 0.66                 & 0.37                & \textbf{0.12}              & 0.05             
\end{tabular}
\vspace{-0.05cm}
\caption{\gls{rmse} of the inclination followed by the end-effector ($\bm{\downarrow}$). }
\label{t:real_evaluation_slope}
\end{subtable}
\vspace{0.1cm}

\begin{subtable}{\linewidth}
\centering
\begin{tabular}{ccccccc}
\hline
\multirow{2}{*}{\textbf{Tool}} & \multicolumn{2}{c}{\textbf{Demo only}} & \multicolumn{2}{c}{\textbf{No FT}} & \multicolumn{2}{c}{\textbf{Fintuned}} \\
                               & x                  & $\sigma$                 & x                    & $\sigma$                   & x                 & $\sigma$                 \\ \hline\hline
Small brush                    & 0.041              & 0.006             & 0.031                & 0.011               & \textbf{0.014}             & 0.002             \\
Sponge                         & 0.065              & 0.034             & 0.051                & 0.015               & \textbf{0.017}             & 0.004             \\
Wide brush                     & 0.017              & 0.008             & 0.038                & 0.007               & \textbf{0.009}             & 0.004             \\
Broom                          & 0.034              & 0.022             & 0.057                & 0.047               & \textbf{0.020}             & 0.009            
\end{tabular}
\vspace{-0.05cm}
\caption{\gls{rmse} of the normal contact force ($\bm{\downarrow}$). }
\label{t:real_evaluation_force}
\end{subtable}
\vspace{0.1cm}

\begin{subtable}{\linewidth}
\centering
\begin{tabular}{ccccccc}
\hline
\multirow{2}{*}{\textbf{Env.}} & \multicolumn{2}{c}{\textbf{Demo only}} & \multicolumn{2}{c}{\textbf{No FT}} & \multicolumn{2}{c}{\textbf{Fintuned}} \\
                               & x                  & $\sigma$                 & x                    & $\sigma$                   & x                  & $\sigma$                \\ \hline\hline
Step                           & 72.74              & 16.45             & 87.07                & 5.61                & \textbf{93.54}              & 4.01            
\end{tabular}
\vspace{-0.05cm}
\caption{Wiped area ($\bm{\uparrow}$). }
\label{t:real_evaluation_wipedarea}
\end{subtable}
\caption{Comparison of task performance using different tools on "Inclined" and "Steps" surfaces on a physical setup.}
\label{t:real_evaluation}
\vspace{0.2cm}
\end{table}
\setlength{\belowcaptionskip}{0.0cm} 

\paraDraft{(C1) Shared pre-trained model}
\subsection{Does the pre-trained feature extraction module capture transferable contact states across tools?}
~\fref{all_tool_pca} illustrates a \gls{pca} applied to a snapshot of the latent space with cell states $\mathbf{c_t} \in \mathbb{R}^{100}$, where $t=5s$, when following a surface with unseen inclinations of $-0.25$, $0.0$, and $0.25$ radians using the simulated brush used for pre-training and four tools used in the real-world environment.
While the first principal component (PC1) delineates the different tools as indicated by the shape of the markers, the second principal component (PC2) demonstrates the separation according to the inclination of the surface, as indicated by the colour of the markers. 
The self-organisation of the latent space based on surface inclination, even across different tools, validates our hypothesis that the pre-trained model captures the contact dynamics between the end-effector, tools, and the environment in a transferable manner across different tools.  
This transferability enables successful execution of downstream tasks with minimal demonstrations when adapting to new tools.
Despite pre-training with a single tool, the latent space captures both environments and tools, which we hypothesise emerged from variations in grasp location and contact forces experienced during pre-training. 

\begin{figure}[t]
\vspace{0.1cm}
\centering
\includegraphics[keepaspectratio, scale=0.28]{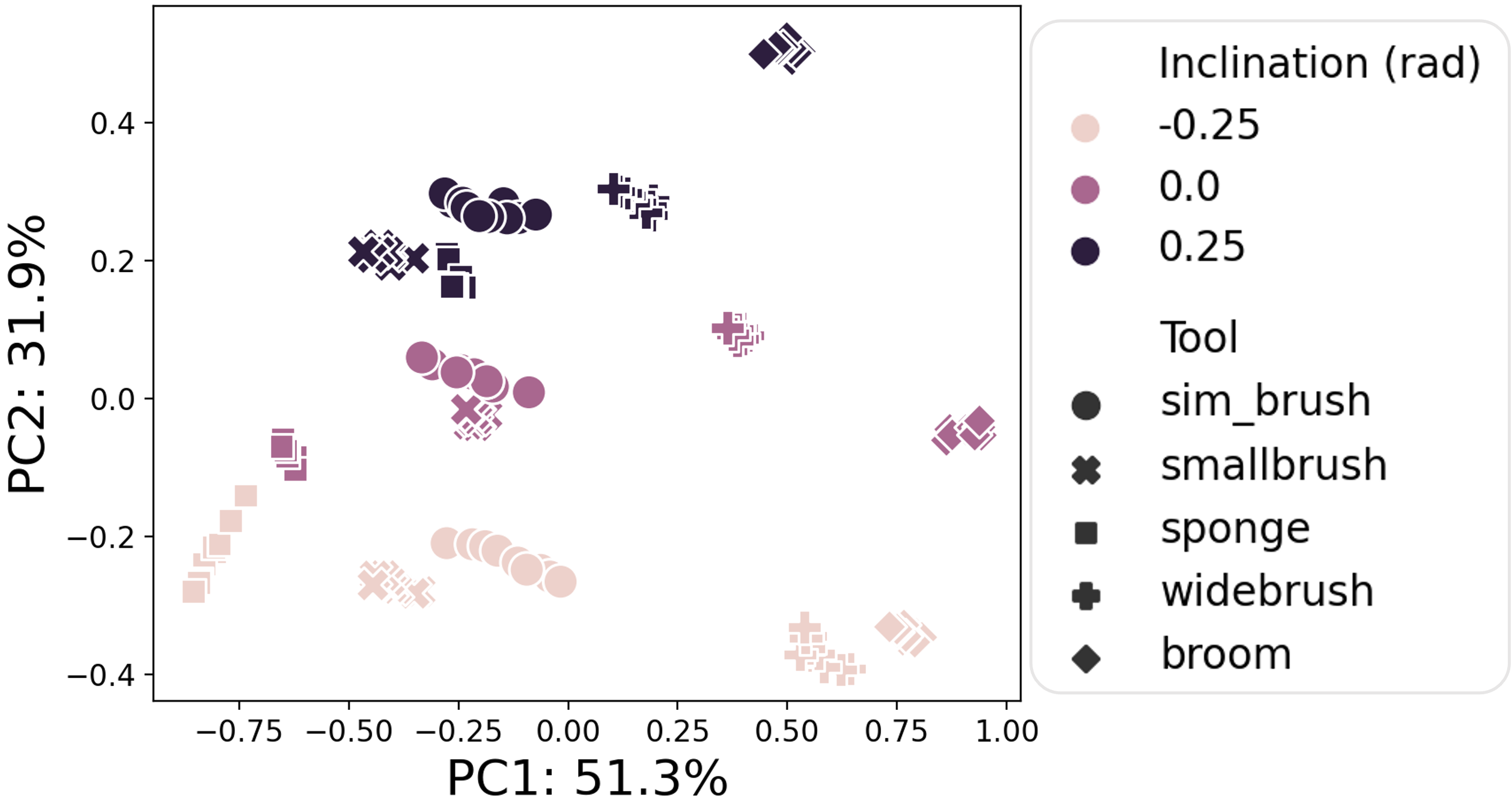}
\vspace{-0.1cm}
\caption{
\gls{pca} applied to a snapshot of the latent space while following a surface with unseen inclinations. 
}
\label{f:all_tool_pca}
\end{figure}


\paraDraft{(C2) Ablation study on multi-modality contribution}
\subsection{How do proximity and tactile sensing contribute to tool manipulation?} 
We investigated the contribution of each sensor modality to surface-following tasks in both simulated and real-world environments, as summarised in~\fref{modality_contribution}. 

\paraDraft{Simulated experiments}
In simulation, all modalities perform similarly in following the desired inclination. 
However, proximity sensing alone struggles to achieve the desired force on inclined surfaces and to detect tool-tip contact on stairs, where tactile information is essential. 
Additionally, the sensor configurations capturing only intrinsic contact, as shown in~\fref{sensor_config}(a)(1)-(3) performed poorly across tasks, as indicated in~\fref{sensor_config}(b)–(d). 
Combining an array of proximity sensors and shear force from tactile sensors which allows to capture both intrinsic and extrinsic contact (\fref{sensor_config}(a)(4)) resulted in the lowest \gls{rmse} of contact force on inclined surfaces and the largest wiped area on stairs. 
These results underscore the importance of our sensor configurations and learning framework combining both sensing to implicitly capture both intrinsic and extrinsic contact for achieving the desired tool-environment contact. 


\begin{figure}[t]
    \centering
    \includegraphics[width=0.9\linewidth]{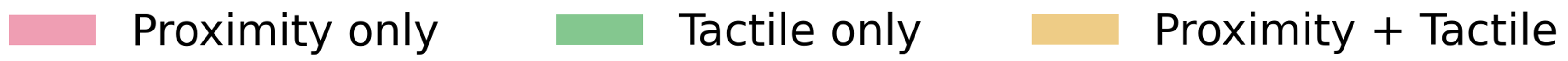}
    
    \vspace{0.05cm} 
    
    \subcaptionbox{Inclination $\bm{\downarrow}$ \label{f:modality_inclination}}{\includegraphics[width=0.33\linewidth]{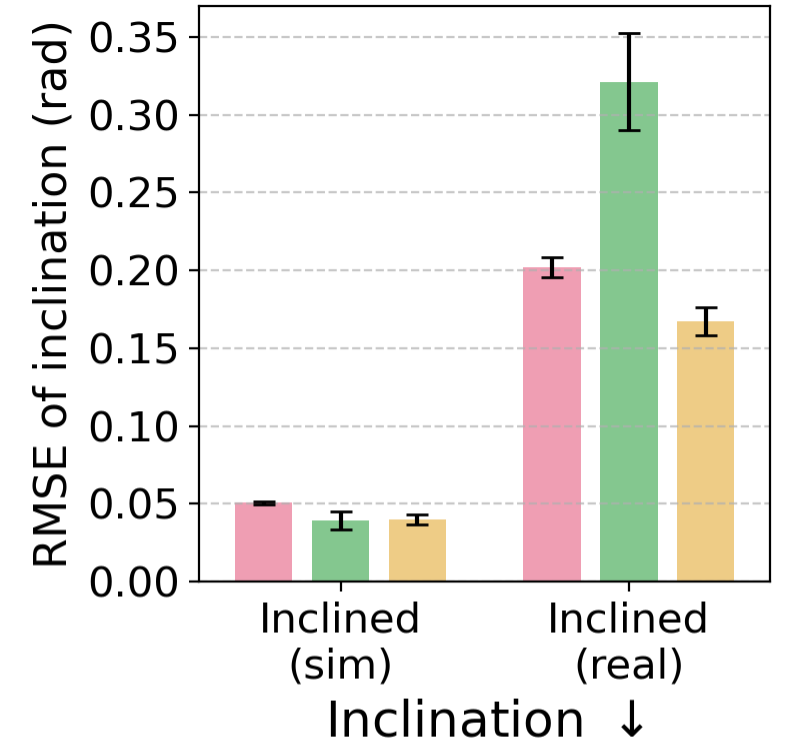}}\hfill
    \subcaptionbox{Normal force $\forcez$ $\bm{\downarrow}$\label{f:modality_force}}{\includegraphics[width=0.33\linewidth]{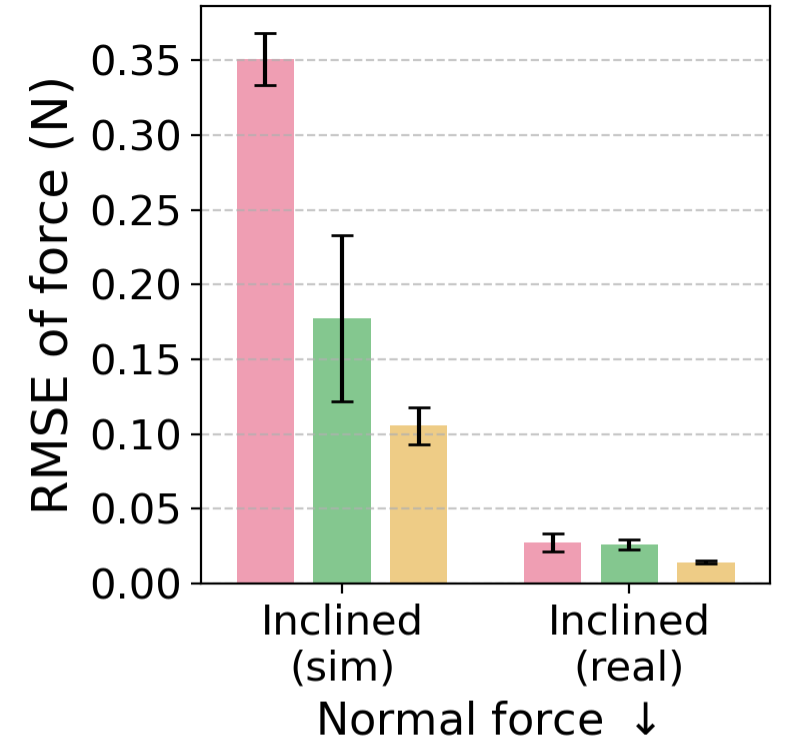}}\hfill
    \subcaptionbox{Wiped area $\bm{\uparrow}$\label{f:modality_wiped}}{\includegraphics[width=0.33\linewidth]{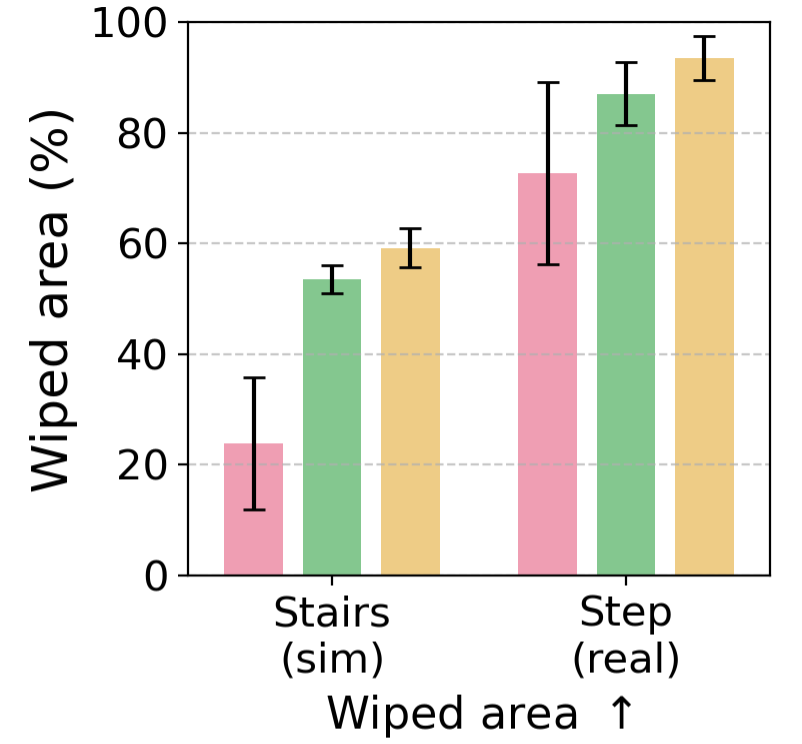}}
    \vspace{0.05cm}
    \caption{Contribution of each modality. }
    \label{f:modality_contribution}
\end{figure}

\begin{figure}[t]
    \centering
    \captionsetup[sub]{font=footnotesize}

    \subcaptionbox{\centering Configurations.\label{f:sensor_config}}[0.23\linewidth]{
        \includegraphics[width=\linewidth]{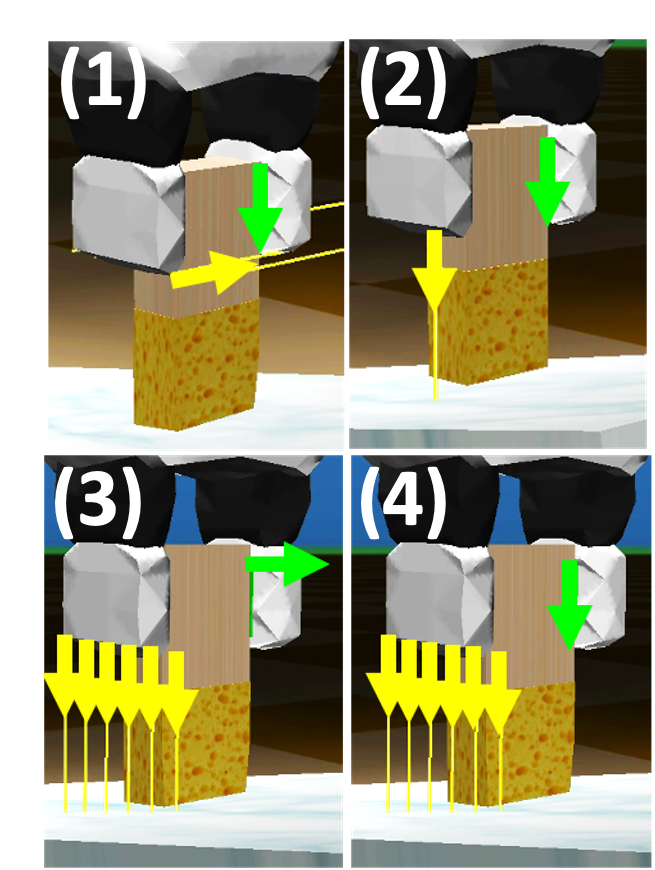}
    }\hfill
    \subcaptionbox{\centering \shortstack{Inclination \\ $\bm{\downarrow}$}\label{f:sensor_inclination}}[0.24\linewidth]{
        \includegraphics[width=\linewidth]{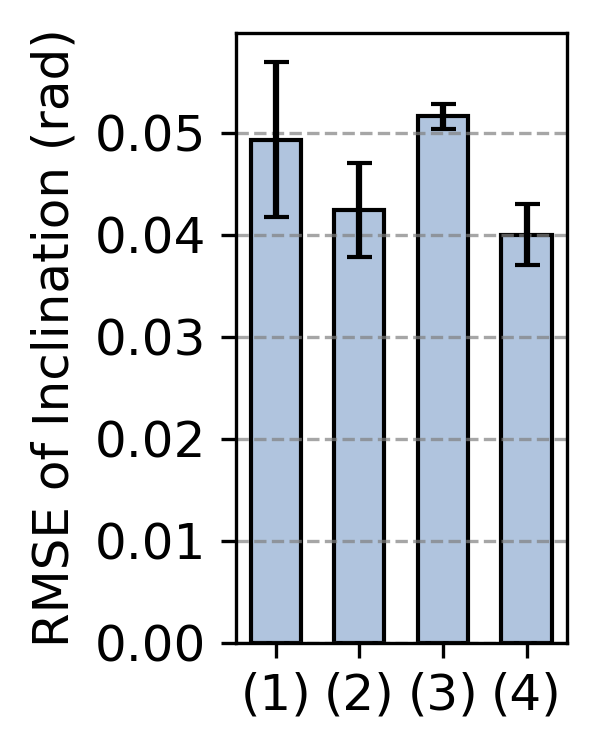}
    }\hfill
    \subcaptionbox{\centering \shortstack{Normal force \\ $\forcez$ $\bm{\downarrow}$}\label{f:sensor_force}}[0.24\linewidth]{
        \includegraphics[width=\linewidth]{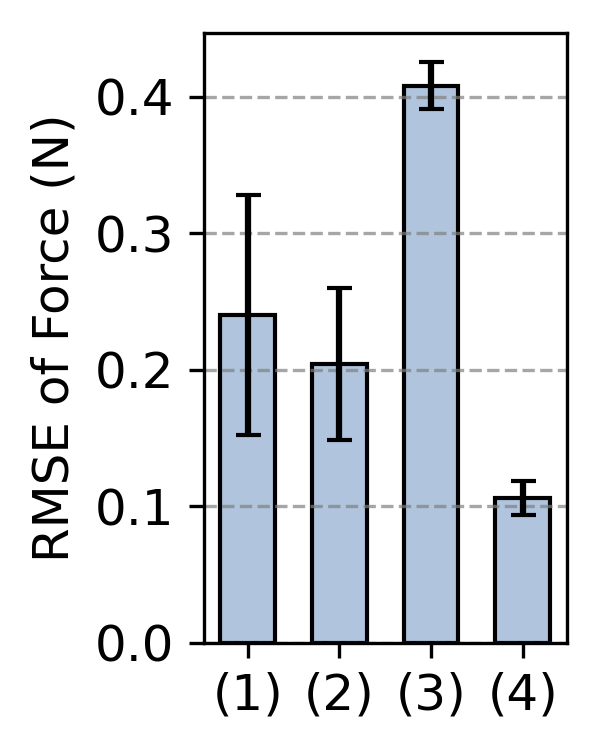}
    }\hfill
    \subcaptionbox{\centering \shortstack{Wiped area \\ $\bm{\uparrow}$}\label{f:sensor_wiped}}[0.24\linewidth]{
        \includegraphics[width=\linewidth]{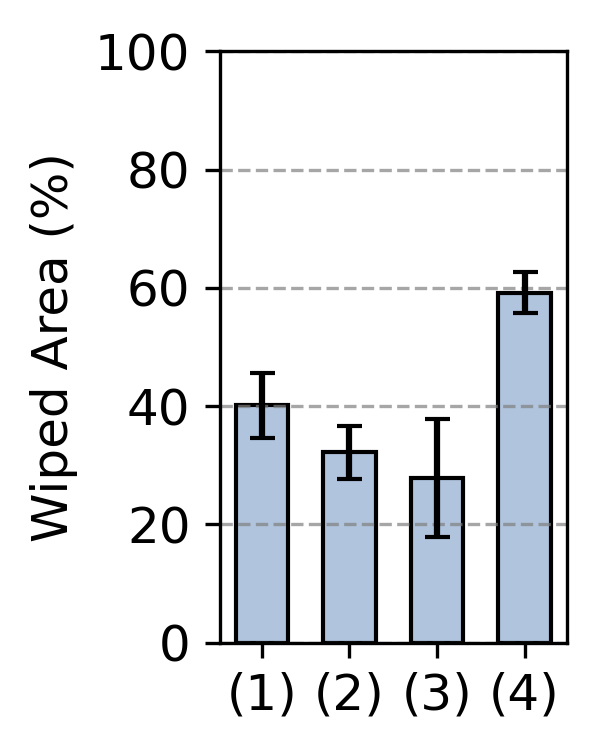}
    }

    \caption{Comparison of different sensor configurations: (1) proximity sensors inside the finger, (2) a single proximity sensor, (3) normal force only, (4) proposed shear and slip force with an array of proximity sensors.}
    \label{f:sensor_config}
    \vspace{0.2cm}
\end{figure}


\paraDraft{Real experiments} 
In real-world environments, combining proximity and tactile sensing significantly improved task performance in both inclined and step environments compared to using either sensor alone. 
While tactile sensing outperformed proximity sensing in simulation, their performances were closer in real-world scenarios, suggesting a larger sim-to-real gap and higher sensor noise for tactile sensors in physical environments.

\paraDraft{(C3) Qualitative experiment on more variations of tools and tasks}
\subsection{How adaptive and robust are the learnt tool-use skills?} 
\paraDraft{Online adaptation } 
Finally, we qualitatively evaluated the online adaptability of the learnt tool-use policy to changing environments. 
We apply t-SNE to the LSTM encoder’s cell state $\mathbf{c_t} \in \mathbb{R}^{100}$ at each timestep $t$ for both training and testing data, with the fixed hyperparameters: perplexity = 30, learning rate = 200, number of iterations = 1000, and using Euclidean distance as the metric. 
~\fref{pca_transition}(1) and~\fref{pca_transition}(3) show the t-SNE-transformed cell states, with grey dashed lines illustrating the transition of cell states in response to changing surface inclination from -0.25 radians (down) to 0.0 radians (flat) and to 0.25 radians (up) in simulation and real settings, respectively. 
Similarly, the model updated cell states when detecting tool-tip contact with walls in the stairs environment in simulation (see~\fref{pca_transition}(2)) and in the step environment in physical setups (see~\fref{pca_transition}(4)). 
This analysis, which shows that our policy can quickly update the contact state encoded in the model’s cell states based on sensing feedback, supports our quantitative results on successful adaptation when hitting the wall in the "Stairs" and "Step" environments in~\tref{sim_rmse_table} and ~\tref{real_evaluation}, as well as online adaptation to changing inclination of the surface shown in~\tref{online_evaluation}. 
Furthermore, when the surface deformed as the robot wiped the paper, as shown in the rightmost target environment in~\fref{experiment_setup}, the robot successfully adapted its motion to the deforming surface. 

\begin{table}[t]
\centering
\scriptsize 
\setlength{\tabcolsep}{3pt} 
\begin{tabular}{c|c@{\hskip 3pt}c@{\hskip 3pt}c|c@{\hskip 3pt}c@{\hskip 3pt}c}
\hline
\textbf{Evaluation} & \multicolumn{3}{c|}{\textbf{Simulation}} & \multicolumn{3}{c}{\textbf{Physical setup}} \\
\textbf{Metric} & 
\makecell[c]{\textbf{Demo only}} & 
\makecell[c]{\textbf{No FT}} & 
\makecell[c]{\textbf{Finetuned}} & 
\makecell[c]{\textbf{Demo only}} & 
\makecell[c]{\textbf{No FT}} & 
\makecell[c]{\textbf{Finetuned}} \\
\hline\hline
\makecell[c]{RMSE of\\slope (rad) $\bm{\downarrow}$} 
  & \makecell{0.082 \\ \scriptsize$\pm$ 0.007} 
  & \makecell{0.052 \\ \scriptsize$\pm$ 0.005} 
  & \makecell{\textbf{0.042} \\ \scriptsize$\pm$ 0.006} 
  & \makecell{0.76 \\ \scriptsize$\pm$ 0.21} 
  & \makecell{0.30 \\ \scriptsize$\pm$ 0.10} 
  & \makecell{\textbf{0.22} \\ \scriptsize$\pm$ 0.05} \\
\makecell[c]{RMSE of\\force (N) $\bm{\downarrow}$} 
  & \makecell{0.72 \\ \scriptsize$\pm$ 0.05} 
  & \makecell{0.29 \\ \scriptsize$\pm$ 0.05} 
  & \makecell{\textbf{0.14} \\ \scriptsize$\pm$ 0.03} 
  & \makecell{0.052 \\ \scriptsize$\pm$ 0.006} 
  & \makecell{0.035 \\ \scriptsize$\pm$ 0.002} 
  & \makecell{\textbf{0.017} \\ \scriptsize$\pm$ 0.004} \\
\hline
\end{tabular}
\caption{
Task performance on online changing inclination from -0.25 rad. (down) to 0.0 rad. (flat) and to 0.25 rad. (up). 
}
\label{t:online_evaluation}
\vspace{0.1cm}
\end{table}

\begin{figure}[t]
\vspace{0.1cm}
\centering
\includegraphics[keepaspectratio, scale=0.55]{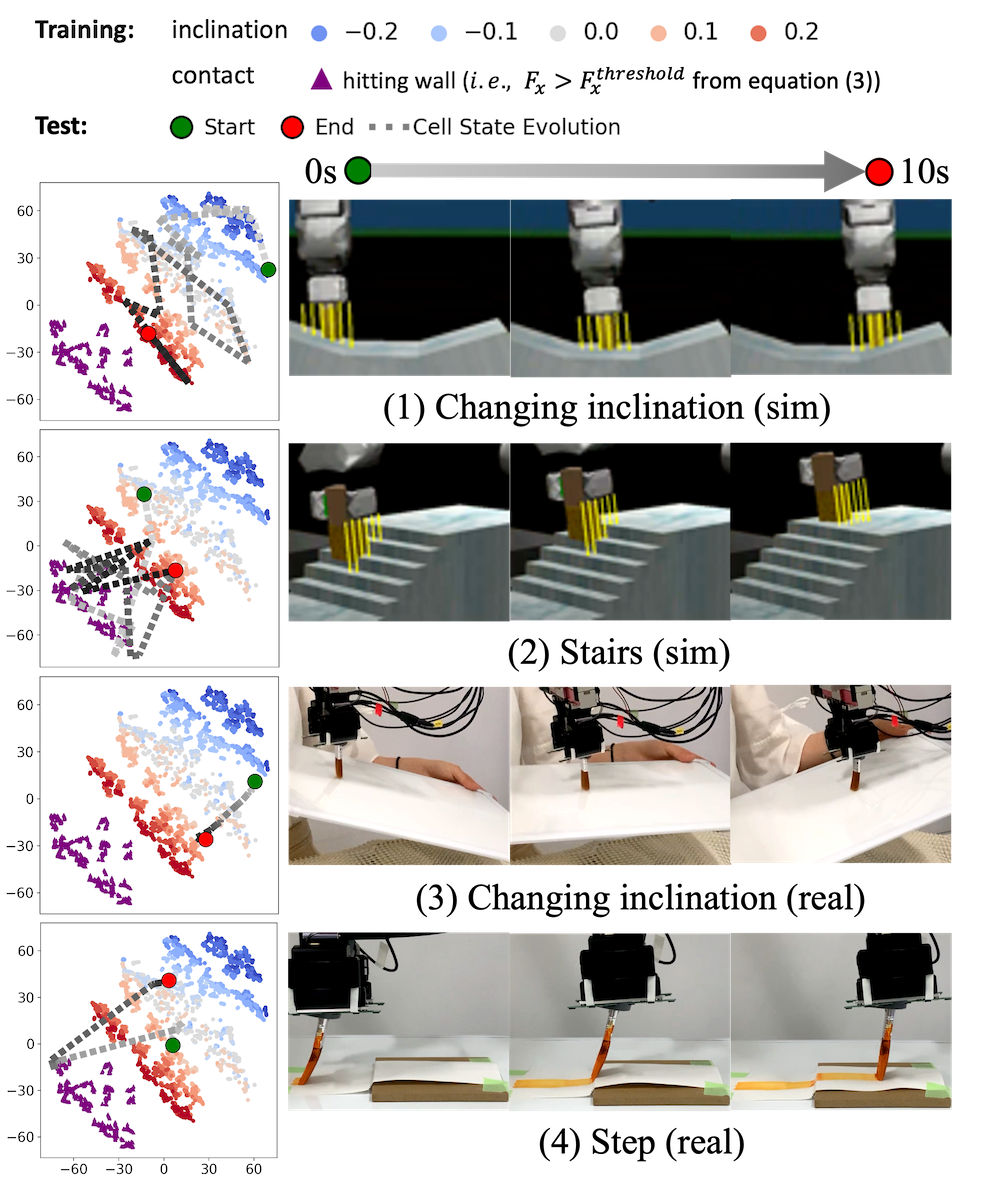}
\caption{
The t-SNE visualisation of cell state evolution in the feature extraction module over time and the adaptation of robot motion on the fly in simulated and physical setups.
}
\setlength\intextsep{0pt}
\label{f:pca_transition}
\end{figure}

\section{CONCLUSIONS}
\label{s:conclusion}


\paraDraft{Summary} 
Our research addresses the challenge of few-shot tool-use skill transfer, enabling robots to manipulate various tools in real environments. 
Our proposed framework leverages inexpensive simulation data to pre-train a base policy, followed by fine-tuning with human demonstrations to bridge the domain gap. 
Experimental results demonstrate that our approach enables robots to manipulate new tools with varying physical properties and adapt to environments with different geometries, with limited demonstrations in surface-following tasks.
Additionally, our analysis indicates that the pre-trained policy's ability to recognise tool-environment contact relationships using proximity and tactile sensing transfers to various downstream tool-use skills.

\paraDraft{Limitations}
In this work, we used only one tool during pre-training; future work will explore multiple tools for more thorough validation of tool generalisation. 
One area of interest is improving robustness to tool slippage by incorporating grip force control or integrating with other grasp mechanisms~\cite{Narita_Nagakari_Conus_Tsuboi_Nagasaka_ICRA2020_naritaslip}. 
Additionally, integrating vision could provide a global view of the environment, complementing local proximity and tactile sensing. 
Finally, applying our method to more complex tasks, such as drawing and spreading butter, offers a promising direction for demonstrating its broader adaptability. 

\addtolength{\textheight}{-12cm}   

\vspace{-0.05cm}
\footnotesize
\bibliographystyle{IEEEtran}
\bibliography{IEEEabrv,references}
\raggedbottom

\end{document}